\documentclass[runningheads]{llncs}
\usepackage[T1]{fontenc}
\usepackage{graphicx}
\usepackage{booktabs}
\usepackage{tabularx}
\usepackage{caption}
\usepackage[hidelinks]{hyperref}

\usepackage{color}

\begin{document}
\newcommand{\OUT}[1]{}

\OUT{\title{Capturing the Whole Team: Efficient Sentence-Level Clinical Provenance Categorization Towards Multidisciplinary NICU Summarization}}
\title{Towards Multidisciplinary Summarization of Hospital Stays: Efficient Sentence-Level Clinical Provenance Categorization}
\titlerunning{Clinical Provenance Modeling for Multidisciplinary NICU Summarization}
%
%
\author{
Baris Karacan\inst{1} \and
Vaibhav Bhargava\inst{1} \and
Barbara Di Eugenio\inst{1} \and
Natalie Parde\inst{1} \and
Catherine Craven\inst{3} \and
Karen Dunn Lopez\inst{2} \and
Andrew Boyd\inst{1} \and
Clinical Data Team\inst{1,2,3}
\thanks{\scriptsize Mary Khetani, Yu-Shan Tseng, Vanessa Barbosa, Julie Vignato, Lindsey Knake, Rajashree Dahal, Emily Spellman, Danielle Hitzel, Janine Petitgout, Kristi Haughey, Amanda Karstens, Brianna Clarahan, Rachel Dawson, Lauren Boyd, Mackenzie Weis, Angie Tipton, Jaewon Bae}
}

\authorrunning{B. Karacan et al.}
%
\institute{
University of Illinois Chicago \and
University of Iowa \and
University of Missouri-Columbia}
%
\maketitle              

\begin{abstract}
Effective "all-team" summarization in high-complexity settings like the Neonatal Intensive Care Unit (NICU) requires aggregating insights from diverse disciplines (physicians, nurses, therapists) spread across hundreds of clinical free-text notes. Simply pooling heterogeneous text often leads to incoherent outputs. Structured summarization therefore first requires accurate categorization of sentence-level provenance across multi-source notes. This pilot study introduces a clinical provenance categorization pipeline using supervised fine-tuning (SFT) of large language models (LLMs). We adapted two Llama-3 models (8B and 70B) to MedSecId, a corpus of 2,002 MIMIC-III (Adult ICU) notes annotated with clinical provenance headers, achieving in-domain Macro F1 scores above 92\% for both models. To evaluate cross-domain generalization, we assessed model capacity (8B vs. 70B) and quantization on a gold-standard dataset of 227 sentence-level spans derived from three multi-disciplinary NICU summaries. Experimental results demonstrate a scale-dependent transfer effect: while SFT produced only marginal changes for the 8B model, it substantially improved the 70B model, increasing Macro F1 by 7\%. Notably, the quantized fine-tuned 70B model outperformed its full-precision baseline while substantially reducing computational requirements. These findings suggest that sufficient model capacity is critical for preserving semantic flexibility during cross-domain clinical transfer and that efficient quantized adaptation can enable structured provenance modeling for downstream summarization.

\end{abstract}

\section{Introduction}
\vspace{-3pt}

Critical care in high-complexity environments like the Neonatal Intensive Care Unit (NICU) relies on synchronized expertise across diverse disciplines, including physicians, nurses, and therapists.  This collaborative model produces unbalanced Electronic Health Records (EHR), where extensive physician documentation can overshadow lower-frequency contributions from specialized services. Standard summarization approaches risk losing these nuanced signals by "pooling" heterogeneous text. To address this challenge, we propose a provenance-aware retrieval pipeline that categorizes sentence-level spans from multi-source clinical notes, preserving multi-disciplinary structure for downstream summarization. 

Clinical section header identification has evolved from rule-based matching \cite{denny2008development} to context-aware supervised language models \cite{zhang2022section,saleh2024tocbert}. However, supervised approaches typically require extensive domain-specific annotation and often struggle to generalize to specialized clinical settings \cite{karacan2026bridging}. To address this limitation, we investigate whether large language models (LLMs) can transfer structured provenance knowledge across domains. Specifically, we evaluate two Llama-3 models \cite{dubey2024llama} (8B and 70B), fine-tuned on adult ICU data, for cross-domain generalization to neonatal care. Our findings reveal a scale-dependent transfer pattern: while both models perform comparably in-domain, the quantized 70B model more effectively preserves structured provenance constraints under domain shift, achieving more balanced cross-domain performance compared to the smaller variant.
\section{Methodology}
\vspace{-5pt}
\subsection{Datasets and Clinical Provenance Schema}
To investigate the transferability of our models, we utilized two distinct clinical datasets reflecting a domain shift from adult to neonatal intensive care.

\begin{itemize}
\item\textbf{Source Domain (Adult ICU):} We leveraged MedSecId~\cite{landes2022new}, comprising 2,002 MIMIC-III notes with annotated headers. To prioritize cross-domain generalization, we distilled the annotations into 17 high-prevalence categories (e.g.,\textit{History of Present Illness, Hospital Course}), resulting in a final corpus of 1,562 notes partitioned into 80/10/10 splits. This curation minimizes overfitting to facility-specific headers, ensuring the model learns robust, transferable clinical structures. To facilitate context-aware provenance modeling, we formulated this data as a structured chat-based instruction-tuning task consisting of 134,640 sentence-level spans (107,777 train, 13,876 val, 12,987 test). Each training instance was defined by a specific system instruction (\textit{"You must assign exactly one section header..."}), a user input containing the target sentence flanked by its immediate predecessor and successor for context, and the expected assistant output corresponding to the ground-truth provenance label.
\item\textbf{Target Domain (NICU):} To evaluate zero-shot transfer, we constructed a NICU Pilot Evaluation Set consisting of 227 sentence-level spans derived from three de-identified, multidisciplinary “all-team” discharge summaries. These summaries were manually annotated by clinical experts to establish a gold standard. The NICU schema comprises 25 fine-grained provenance categories spanning general discharge sections (e.g., \textit{Brief Admission Summary}), discipline summaries (\textit{PT, OT, SLP}), care coordination domains (\textit{Feeding-Nutrition, Family-Social}) and an organ-system decomposition of the hospital course (e.g., \textit{Respiratory, Cardiovascular}). This representation only partially overlaps with adult ICU headers and introduces substantially greater granularity, providing a stringent test of semantic generalization under schema shift. The study was approved by the University of Illinois Chicago IRB. 
\end{itemize}

\subsection{Model Architectures and Efficient Fine-Tuning}
We fine-tuned Llama-3.1-8B-Instruct and Llama-3.3-70B-Instruct models using QLoRA~\cite{dettmers2023qlora}, reducing model weight memory by approximately 75\% relative to 16-bit storage. This enabled a full 70B model replica to fit within the 80GB memory budget of an A100 GPU, allowing data-parallel training across four GPUs. Both models were quantized to 4-bit Normal Float (NF4) precision, with LoRA adapters~\cite{hu2022lora} ($r=16, \alpha=32$) attached to all linear projections, including MLP \textit{gate}, \textit{up}, and \textit{down} layers, rather than restricting adaptation to attention modules. This substantially increased trainable parameter coverage and allowed modification of both attention patterns and feed-forward representations. Following standard scaling practice, we used a learning rate of $2e^{-4}$ for the 8B model and $1e^{-4}$ for the 70B model. Both models converged stably with minimal train–validation loss gaps ($<0.08$), suggesting effective generalization without evidence of overfitting.

\section{Results}
\vspace{-5pt}
\subsection{In-Domain Performance (MedSecId)}
Before evaluating cross-domain transfer, we first assessed the instruction-tuning pipeline on the source domain. Both architectures demonstrated strong alignment with the clinical provenance schema. As shown in Table~\ref{tab:indomain_results}, the 8B and 70B models achieved nearly identical weighted F1-scores of 0.94 on the held-out test set ($N=12{,}987$ spans). Importantly, Macro F1-scores (0.92–0.93) closely matched the weighted metrics, indicating that performance was consistent across both high- and low-frequency categories rather than being driven primarily by prevalent classes. For example, the low-frequency provenance \textit{Review of Systems} ($N=56$ spans; 0.43\% of the test set) achieved F1-scores between 0.81 and 0.84, demonstrating meaningful generalization even under limited class support. The close agreement between weighted and macro averages establishes a robust in-domain baseline. Notably, the larger 70B model did not exhibit a statistically meaningful advantage over the 8B variant under domain-specific supervision, suggesting that parameter scale alone does not substantially improve schema internalization in the source setting.

\begin{table*}[t] 
\centering
\caption{\footnotesize{In-Domain Evaluation on MedSecId Test Set ($N=12,987$ spans)}}
\label{tab:indomain_results}
\footnotesize
\begin{tabularx}{\textwidth}{l*{6}{>{\centering\arraybackslash}X}}
\toprule
\textbf{Model} & \multicolumn{3}{c}{\textbf{Macro Average}} & \multicolumn{3}{c}{\textbf{Weighted Average}} \\
\cmidrule(lr){2-4} \cmidrule(lr){5-7}
& \textbf{Precision} & \textbf{Recall} & \textbf{F1} & \textbf{Precision} & \textbf{Recall} & \textbf{F1} \\
\midrule
Llama-3.1-8B & \textbf{0.94} & \textbf{0.92} & 0.92 & \textbf{0.94} & \textbf{0.94} & \textbf{0.94} \\
Llama-3.3-70B & \textbf{0.94} & \textbf{0.92} & \textbf{0.93} & \textbf{0.94} & \textbf{0.94} & \textbf{0.94} \\
\bottomrule
\end{tabularx}
\end{table*}

\subsection{Cross-Domain Transfer to NICU}
Evaluation on the NICU Pilot Set ($N=227$ spans) revealed a distinct divergence in generalization capabilities (Table~\ref{tab:nicu_results}). Although both architectures achieved comparable in-domain performance (0.94 weighted F1), their ability to transfer structured provenance supervision to neonatal care differed meaningfully. Llama-3.3-70B (SFT) achieved the strongest overall performance, reaching a weighted F1 of 0.60, a 4-point improvement over its base version. More notably, its Macro F1 improved from 0.52 to 0.59. This indicates that the 70B model has the parameter capacity to internalize the structured constraints of the clinical schema while maintaining the semantic flexibility required to recognize specialized, low-prevalence NICU provenance categories that it had not seen during fine-tuning. In contrast, Llama-3.1-8B demonstrated only modest transfer gains. While weighted F1 improved slightly from 0.47 to 0.50, Macro F1 remained essentially unchanged (0.53 to 0.52). This near-parity suggests that fine-tuning primarily reinforced performance on higher-frequency classes without meaningfully improving minority-category generalization. Unlike the 70B model, the 8B architecture appears constrained in its ability to translate structured supervision into robust cross-domain improvements. These results suggest that larger-capacity models are more effective at translating structured supervision into balanced cross-domain performance.



\begin{table*}[t]
\centering
\caption{\footnotesize{Cross-Domain Evaluation on NICU Pilot Set ($N=227$ spans).}}
\label{tab:nicu_results}
\footnotesize
\setlength{\tabcolsep}{3.5pt} 
\renewcommand{\arraystretch}{0.95} 
\begin{tabularx}{\textwidth}{ll*{6}{>{\centering\arraybackslash}X}}
\toprule
\textbf{Model} & \textbf{Version} & \multicolumn{3}{c}{\textbf{Macro Average}} & \multicolumn{3}{c}{\textbf{Weighted Average}} \\
\cmidrule(lr){3-5} \cmidrule(lr){6-8}
& & \textbf{Precision} & \textbf{Recall} & \textbf{F1} & \textbf{Precision} & \textbf{Recall} & \textbf{F1} \\
\midrule
Llama-3.1-8B & Base & 0.54 & \textbf{0.62} & \textbf{0.53} & 0.53 & 0.50 & 0.47 \\
& SFT & \textbf{0.61} & 0.55 & 0.52 & \textbf{0.66} & \textbf{0.52} & \textbf{0.50} \\
\midrule
Llama-3.3-70B & Base & 0.57 & 0.57 & 0.52 & 0.65 & 0.58 & 0.56 \\
& SFT & \textbf{0.64} & \textbf{0.69} & \textbf{0.59} & \textbf{0.70} & \textbf{0.62} & \textbf{0.60} \\
\bottomrule
\end{tabularx}
\end{table*}

\section{Conclusion and Future Work}
\vspace{-5pt}
This study demonstrates that while SFT enables large-scale LLMs to learn complex clinical provenance schemas, parameter scale remains an important determinant of cross-domain robustness. Our results reveal a clear scale-dependent transfer pattern: the 70B model successfully transferred learned structures from adult to neonatal care, whereas the 8B model exhibited only marginal cross-domain gains. A primary limitation of this work is the composition of the NICU Pilot Set, which consists of 227 sentence-level spans derived from three multi-disciplinary discharge summaries. While the performance differences are numerically meaningful, a McNemar exact test yielded a p-value of 0.24. This lack of statistical significance is likely due to the limited sample size and high intra-note correlation. Notably, developing such gold-standard sets is exceptionally resource-intensive; verifying provenance in "all-team" summaries requires manually cross-referencing a broad range of discipline-specific progress notes to ensure ground-truth accuracy. This logistical complexity inherently constrains the volume of annotated data available for specialized neonatal evaluation. 

Future work will prioritize expanding the NICU evaluation set to include a broader range of "all-team" summaries. We plan to investigate how these provenance classifiers can serve as retriever backbones for downstream clinical summarization. By accurately segmenting notes into discipline-specific zones (e.g., \textit{Feeding-Nutrition, Caregiver Education}), we can provide LLM-based summarizers with more precise contextual grounding, potentially reducing hallucinations and improving the clinical utility of generated summaries. Importantly, the use of QLoRA-based 4-bit quantization enabled efficient adaptation of the 70B model within a single 80GB GPU memory budget, demonstrating that large-capacity provenance modeling can remain computationally feasible in clinical research settings. Overall, our findings suggest that for cross-domain clinical transfer, larger-capacity models provide more consistent and balanced gains, while smaller models yield comparatively limited improvements.

\end{document}